\documentclass[conference]{IEEEtran}
\IEEEoverridecommandlockouts    

\usepackage{multirow}
\usepackage{lipsum}
\usepackage{amsmath}
\usepackage{amssymb}
\usepackage[ruled,vlined]{algorithm2e}
\usepackage{graphicx}
\graphicspath{{./Figures/}}
\usepackage[hidelinks]{hyperref}
\usepackage{amsfonts}
\usepackage{algorithmic}
\usepackage{array}
\usepackage{booktabs}
\usepackage[caption=false,font=normalsize,labelfont=sf,textfont=sf]{subfig}
\usepackage{textcomp}
\usepackage{stfloats}
\usepackage{cleveref}
\usepackage{url}
\usepackage{verbatim}
\usepackage{cite}
\usepackage{xcolor}
\usepackage{makecell}
\usepackage{subcaption}

\title{\LARGE \bf Damper-B-PINN: Damper Characteristics-Based Bayesian Physics-Informed Neural Network for Vehicle State Estimation}

 \author{
 	\parbox{\textwidth}{%
 		\centering
 		Tianyi Zeng$^{1*}$, Tianyi Wang$^{2*\dag}$, Zimo Zeng$^{3}$, Feiyang Zhang$^{4}$, Jiseop Byeon$^{2}$, Yujin Wang$^{4}$, Yajie Zou$^{5}$, Yangyang Wang$^{4}$, Junfeng Jiao$^{6}$, Christian Claudel$^{2}$, Xinbo Chen$^{4}$%
 	}%
 	\thanks{$^\dag$Corresponding author: Tianyi Wang.}%
    \thanks{$^{*}$These authors contributed equally to this work.}%
    \thanks{$^{1}$School of Automation and Intelligent Sensing, Shanghai Jiao Tong University, Shanghai 200240, China.
 		{\tt\small zengtianyi@sjtu.edu.cn}}%
 	\thanks{$^{2}$Department of Civil, Architectural, and Environmental Engineering, The University of Texas at Austin, Austin, TX 78712, USA.
 		{\tt\small bonny.wang@utexas.edu, jsbyeon@utexas.edu, christian.claudel@utexas.edu}}%
    \thanks{$^{3}$College of Electrical Engineering, Zhejiang University, Hangzhou 310027, China.
 		{\tt\small zimozeng904@gmail.com}}%
    \thanks{$^{4}$School of Automotive Studies, Tongji University, Shanghai 201804, China.
 		{\tt\small 2151166@tongji.edu.cn, 2510180@tongji.edu.cn, wyangyang@tongji.edu.cn, chenxinbo@tongji.edu.cn}}%
    \thanks{$^{5}$Key Laboratory of Road and Traffic Engineering of Ministry of Education, Tongji University, Shanghai 201804, China.
 		{\tt\small yajiezou@hotmail.com}}%
    \thanks{$^{6}$School of Architecture, The University of Texas at Austin, Austin, TX 78712, USA.
 		{\tt\small jjiao@austin.utexas.edu}}%
 }

\begin{document}
	
	\maketitle
	\thispagestyle{empty}
	\pagestyle{empty}
	
	\begin{abstract}
Accurate state estimation is fundamental to intelligent vehicles. 
Wheel load, one of the most important chassis states, serves as an essential input for advanced driver assistance systems (ADAS) and exerts a direct influence on vehicle stability and safety. 
However, wheel load estimation remains challenging due to the complexity of chassis modeling and the susceptibility of nonlinear systems to noise. 
To address these issues, this paper first introduces a refined suspension linkage-level modeling approach that constructs a nonlinear instantaneous dynamic model by explicitly considering the complex geometric structure of the suspension. 
Building upon this, we propose a damper characteristics-based Bayesian physics-informed neural network (Damper-B-PINN) framework to estimate dynamic wheel load, which leverages the suspension dynamics as physical guidance of PINN while employing Bayesian inference to mitigate the effects of system noise and uncertainty. 
Moreover, a damper-characteristic physics conditioning (DPC) module is designed for embedding physical prior.
The proposed Damper-B-PINN is evaluated using both high-fidelity simulation datasets generated by CarSim software and real-world datasets collected from a Formula Student race car. 
Experimental results demonstrate that our Damper-B-PINN consistently outperforms existing methods across various test conditions, including extreme ones. 
These findings highlight the potential of the proposed Damper-B-PINN framework to enhance the accuracy and robustness of dynamic wheel load estimation, thereby improving the reliability and safety of ADAS applications.
	\end{abstract}
	

\begin{figure}[t]
    \centering
    \vspace{6pt}
    \includegraphics[width=1\linewidth]{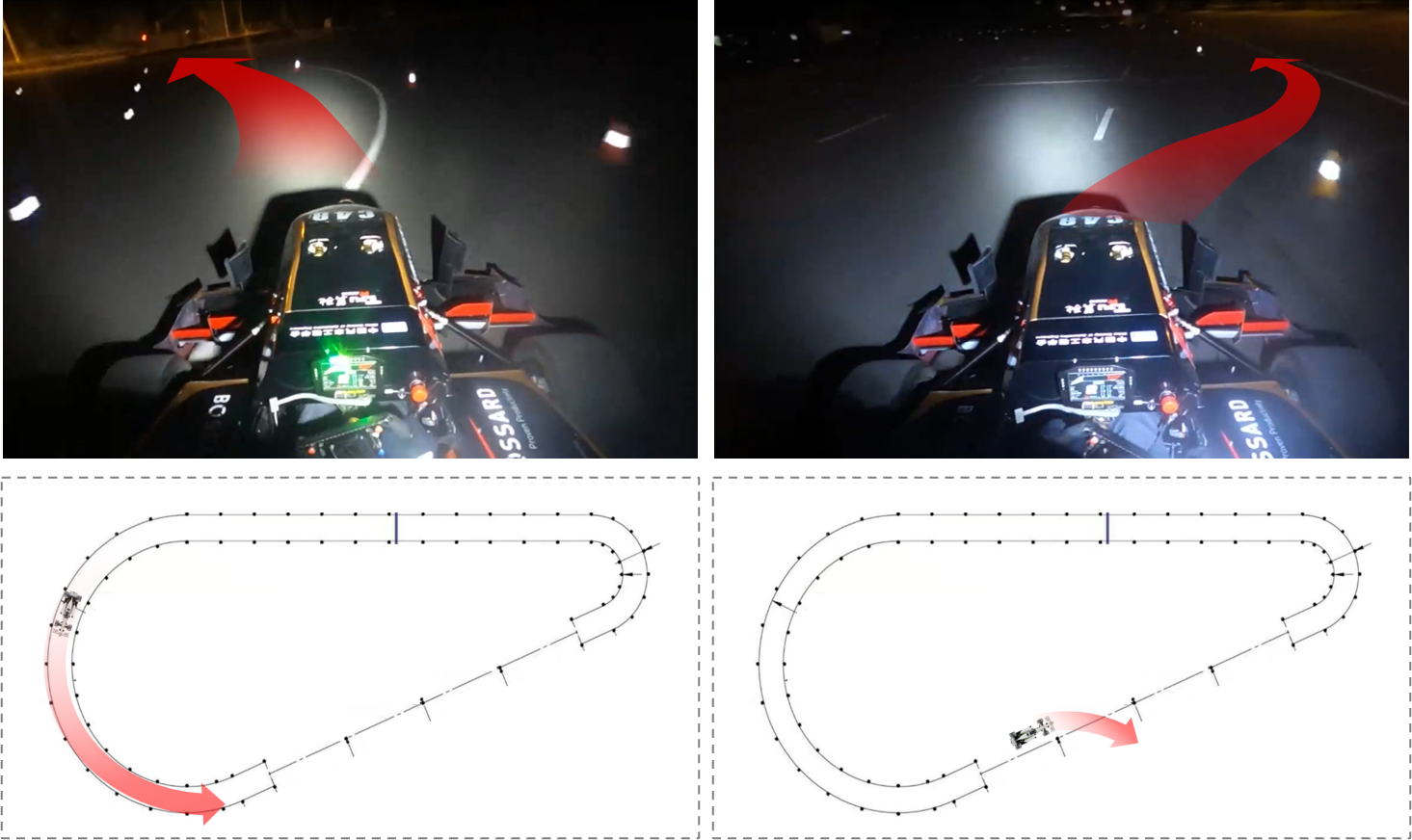}
    \caption{Real-world testing platform (A Formula Student racing car) and standard testing circuits.}
    \label{fig:race}
\end{figure}

\section{Introduction}
\label{introduction}

State estimation in real-world engineering systems subject to noise such as wheel load estimation in vehicle chassis systems, remains a longstanding challenge due to the complex and imperfect nonlinear relationships between inputs and outputs \cite{zeng2024wheel, zeng2024analysis, wang2025hlcg}. 
Traditional methods like EKF sometimes fail to achieve accurate estimation in such real-world scenarios \cite{liu2025one}.
In recent years, data-driven methods have increasingly been employed to approximate closures in nonlinear models and to estimate parameters and functional relationships within mathematical frameworks \cite{wang2024drive, wang2025rad}. 
Deep learning approaches have shown promise for handling such problems, with notable success in solving partial differential equations (PDEs) from sparse or noisy data \cite{Raja2025}. 
Among these, physics-informed neural networks (PINNs) have emerged as a powerful paradigm for both forward (inference) and inverse (identification) problems due to their ability to integrate physical laws into the learning process \cite{Guan2024, Sun2024}. 
By embedding governing equations into the loss function, PINNs can infer unknown parameters and reconstruct solutions from partial observations, making them well-suited for complex state estimation tasks in nonlinear dynamical systems \cite{Long2021}.

Recent studies on PINNs have explored their applications in various complex engineering fields, such as robotics \cite{Yang2023}, automotive \cite{EV-PINN}, and transportation \cite{Long2022}, demonstrating strong performance across diverse tasks.
Additionally, novel optimization methods have been proposed to enhance the accuracy and generalization of PINNs \cite{physics-informed-node}.
Despite the fact that PINNs have demonstrated success in integrating physics into deep neural network (DNN) frameworks, enabling the concurrent use of physics as explicit knowledge and data as implicit knowledge, they lack built-in uncertainty quantification, limiting their applicability in scenarios with high noise levels.

The traditional approach to estimate uncertainty in DNNs relies on Bayes' theorem, as exemplified by Bayesian neural networks (BNNs) \cite{Gal2016-1}. 
While Bayesian inference methods have been developed for quantifying uncertainties in physics problems, most Bayesian approaches require informative prior knowledge about system parameters, which are expected to change frequently for inverter-dominated power systems \cite{Petra2017}.
In contrast to informative priors, weakly informative priors can be generally applied to the whole range of system parameters.
However, Bayesian methods often incur significant additional computational costs, as they need more parameters and longer convergence times.

Recently, physics-informed BNNs have been introduced as a hybrid approach that integrates PINNs and Bayesian techniques to quantify uncertainties in both data and models \cite{MBPINN}.
Considering sparse and noisy measurements in boundary conditions and source terms in real-world applications, Yang et al. \cite{B-PINNs} solved both forward and inverse problems involving linear and nonlinear PDEs with noisy data.

\begin{figure*}[ht]
\centering
\vspace{6pt}
\includegraphics[width=1\linewidth]{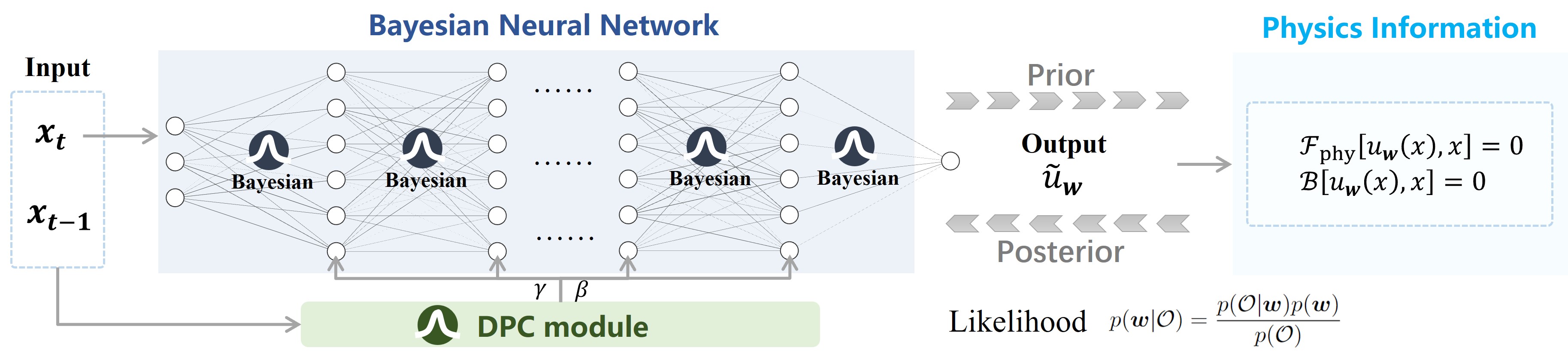}
\caption{Overview of the proposed damper characteristics-based Bayesian physics-informed neural network (Damper-B-PINN) framework: \textbf{(1) Bayesian Estimation:} We employ a novel variational inference method with normal-sigmoid dropout to estimate the posterior distribution of the network parameters. \textbf{(2) Physics Conditioning Embedding:} A damper characteristic feature-wise conditioning module deeply embeds physical information to guide the model.}
\label{frame}
\end{figure*}

In the wheel load estimation task, there are two main challenges: \textbf{(1)} Inaccurate dynamic modeling will mislead the PINN, ultimately resulting in the physical model failing to match the data distribution. \textbf{(2)} The chassis sensors have significant noise, and traditional PINN is only constrained through the physical loss, making it fail to stably embed the physical information during the training process. 
To address these problems, we design a damper characteristics-based Bayesian physics-informed neural network (Damper-B-PINN), a BNN framework inspired by damper properties and introduce a physical model as a priori knowledge within the network structure to enhance accuracy in system estimation, as shown in Figure \ref{frame}. 
The main contributions of this paper are summarized as follows: 
\begin{itemize}
    \item \textbf{A suspension linkage-level modeling method}, which reconstructs the highly nonlinear suspension geometry through refined instantaneous dynamic modeling, improving the reliability of physical information.
    \item \textbf{A Bayesian PINN framework with a damper-characteristic physics conditioning (DPC) module}, which is introduced to reduce the impact of environmental noise on systems with deeply embedded physics information, enhancing the robustness of networks.
    \item \textbf{A comprehensive verification of the framework}, which includes both simulations and real-world Formula Student race car experiments (Figure \ref{fig:race}), demonstrating the effectiveness even under extreme conditions.
\end{itemize}

\section{Related Works}
\label{related_works}

\subsection{Physics-Informed Neural Networks}

The baseline PINN algorithms have achieved remarkable success in the solution of many linear and nonlinear PDEs.
For instance, PINNs have been utilized for modeling cardiovascular flows, constraining the output to satisfy physical conservation laws through one-dimensional models of pulsatile blood flow \cite{Kissas2020}. 
The model proposed in \cite{pinn-energy-ngd} encoded the physical principles of the hydraulic fracturing process, described in the form of PDEs, in DNNs. 
Meng et al. \cite{Meng2023} combined the first-order reliability method with PINNs to avoid calculating the real structure response, which tackled computational difficulty and inefficiency.
Recently, many new PINN architectures, such as conservative PINNs \cite{jagtap2020conservative} and nonlocal PINNs \cite{pang2020npinns}, have been proposed to improve the robustness and efficiency of PINNs.
However, current PINN-based methods often lead to non-convergence issues and low accuracy due to system uncertainty and nonlinearity.
Consequently, the application of PINNs in reliability assessment remains under-explored.

\subsection{Physics-Informed Bayesian Neural Networks}

In the context of power systems, the robustness of physics-informed BNNs against noise-induced uncertainty for system identification has been evaluated, outperforming parsimonious approaches such as sparse identification of nonlinear dynamics \cite{Stock2024}. 
PINNs have also been employed as a surrogate model to solve Bayesian inverse problems, significantly reducing computational time by retaining initial update results to ensure the global nature of the surrogate model \cite{Nabian2020}. 
In another work, an offline-online computational strategy that coupled classical sampling methods with PINN-based approaches for Bayesian inverse problems achieved a substantial reduction in overall computational time while maintaining accuracy \cite{Li2023}. 
Physics-informed BNNs and ensembled PINNs have also been leveraged to quantify uncertainties arising from noisy and incomplete data in governing equations, providing reasonable uncertainty bounds and improving the reliability of predictions \cite{Correcting-PINNs}.
Besides, Li et al. \cite{li2025physics} designed a physics-informed deep learning framework for forward simulation and unknown parameter inference in nondiffusive thermal modeling using sparse and noisy data.
However, existing physics-informed BNN-based models have not been thoroughly validated for complex real-world engineering problems, such as real-time dynamic wheel load estimation.

\section{Methodology}
\label{method}

\subsection{Problem Formulation}

Vehicle dynamic wheel load is crucial data for vehicle chassis design and automatic control. 
The idealized expression of wheel load vector $\boldsymbol{F}_z$ derived from a series of sensor data $\boldsymbol{x}$ is as follows:
\begin{equation}
    \boldsymbol{F}_z=f(\boldsymbol{x}),
\end{equation}
where $f$ is a function or a neural network.

\begin{figure}[ht]
    \centering
    \vspace{6pt}
    \includegraphics[width=1\linewidth]{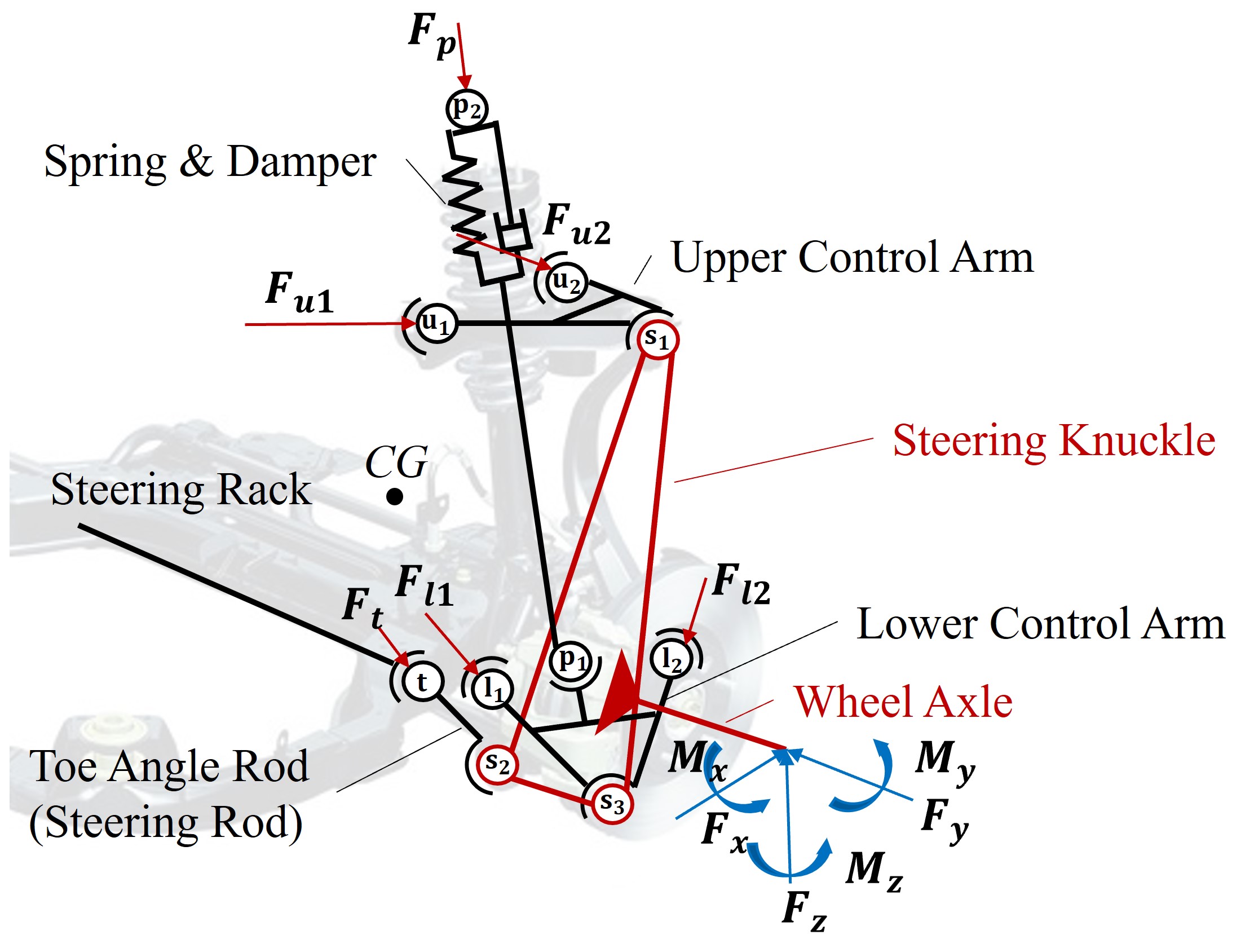}
    \caption{Front suspension model for wheel load estimation: Two-degree-of-freedom (2 DOF) with bumping and steering.}
    \label{fig:sus}
\end{figure}

The estimation of dynamic wheel load presents two fundamental challenges. 
First, accurate estimation is highly dependent on precise chassis modeling, as the wheel–suspension system exhibits strong nonlinearities that are difficult to capture with simplified representations. 
Second, the sensors commonly employed for wheel load estimation are mounted on the chassis and are thus heavily affected by noise and disturbances, which can compromise the reliability of conventional computational methods.
To address these issues, this section first introduces a refined suspension linkage model that reconstructs the nonlinear geometric and mechanical relationships within the suspension system. 
This modeling approach captures the instantaneous dynamic behavior of the suspension with higher fidelity, providing a reliable physical basis for load estimation. 
Building upon this, a computational method based on a physics-informed BNN is presented, which integrates the suspension model as physical prior knowledge within the learning architecture. 
The Bayesian formulation further enables systematic handling of measurement noise and model uncertainty, thereby enhancing both robustness and accuracy in wheel load estimation.
An overview of the proposed Damper-B-PINN framework is illustrated in Figure \ref{frame}, which highlights the integration of physics-based modeling and Bayesian inference within the neural network architecture.

\subsection{Suspension Linkage-Level Modeling}

Since vehicle suspensions typically consist of multiple connecting rods forming complex single-degree-of-freedom (1 DOF) or two-degree-of-freedom (2 DOF) structures, their geometric parameters evolve dynamically during driving \cite{zeng2023dynamic}. 
For computational simplicity, the mass of individual suspension links is neglected, and only the unsprung mass $m_u$, which includes the wheel and steering knuckle, is considered.
The hinge joints of the suspension system, are commonly referred to as ``hard points''. 
Detailed components are labeled in Figure \ref{fig:sus}.
In this study, the hard points are defined as follows:
$\bold{u_1}$ and $\bold{u_2}$ (front and rear points of the upper control arm, UCA);
$\bold{l_1}$ and $\bold{l_2}$ (front and rear points of the lower control arm, LCA);
$\bold{p_1}$ and $\bold{p_2}$ (lower and upper joints of the spring-damper assembly);
$\bold{t}$ (hinge joint between the steering knuckle and steering rack);
and $\bold{s_1}$, $\bold{s_2}$, and $\bold{s_3}$ (upper, front, and lower points of the steering knuckle).

Here, $CG$ denotes the center of mass; $\boldsymbol{F_p}$, $\boldsymbol{F_{u1}}$, $\boldsymbol{F_{u2}}$, $\boldsymbol{F_t}$, $\boldsymbol{F_{l1}}$, $\boldsymbol{F_{l2}}$ represent suspension force vectors; $\boldsymbol{F_x}$, $\boldsymbol{F_y}$, $\boldsymbol{F_z}$ are tire forces transmitted from the wheel; and $\boldsymbol{M_x}$, $\boldsymbol{M_y}$, $\boldsymbol{M_z}$ are moments about the wheel axle. 

We define the vector from $CG$ to hard point $i$ as $\overrightarrow{i}$ (e.g., $\overrightarrow{\bold{u_1}}$) and the direction vector from hard point $i$ to $j$ as $\overrightarrow{ij}$ (e.g., $\overrightarrow{\bold{u_1}\bold{s_1}}$). 
Force vectors can thus be expressed as:
\begin{equation}
\begin{aligned}
\label{eq2}
    &\boldsymbol{F}_p = F_p \cdot \overrightarrow{\bold{p_2}\bold{p_1}}, \quad 
    \boldsymbol{F}_{u1} = F_{u1} \cdot \overrightarrow{\bold{u_1}\bold{s_1}}, \\
    &\boldsymbol{F}_{u2} = F_{u2} \cdot \overrightarrow{\bold{u_2}\bold{s_1}}, \quad 
    \boldsymbol{F}_{t} = F_{t} \cdot \overrightarrow{\bold{t}\bold{s_2}}, \\
    &\boldsymbol{F}_{l_1} = F_{l_1} \cdot \overrightarrow{\bold{l_1}\bold{s_3}}, \quad 
    \boldsymbol{F}_{l_2} = F_{l_2} \cdot \overrightarrow{\bold{l_2}\bold{s_3}},
\end{aligned}
\end{equation}
where unbolded coefficients represent scalar magnitudes of the corresponding forces. 

The suspension force matrix is:
\begin{equation}
    \boldsymbol{F_{\text{sus}}} = \left[ \boldsymbol{F}_p, \boldsymbol{F}_{u_1}, \boldsymbol{F}_{u_2}, \boldsymbol{F}_{t}, \boldsymbol{F}_{l_1}, \boldsymbol{F}_{l_2} \right].
\end{equation}

Moments of these forces about $CG$ are calculated as:
\begin{equation}
\begin{aligned}
    &\boldsymbol{M}_p = \overrightarrow{\bold{p_2}} \times \boldsymbol{F_p}, \quad 
    \boldsymbol{M}_{u_1} = \overrightarrow{\bold{u_1}} \times \boldsymbol{F_{u_1}}, \\
    &\boldsymbol{M}_{u_2} = \overrightarrow{\bold{u_2}} \times \boldsymbol{F_{u_2}}, \quad 
    \boldsymbol{M}_{t} = \overrightarrow{\bold{t}} \times \boldsymbol{F_t}, \\
    &\boldsymbol{M}_{l_1} = \overrightarrow{\bold{l_1}} \times \boldsymbol{F_{l_1}}, \quad 
    \boldsymbol{M}_{l_2} = \overrightarrow{\bold{l_2}} \times \boldsymbol{F_{l_2}}.
\end{aligned}
\end{equation}

The moment matrix is:
\begin{equation}
    \boldsymbol{M_{\text{sus}}} = \left[ \boldsymbol{M}_p, \boldsymbol{M}_{u_1}, \boldsymbol{M}_{u_2}, \boldsymbol{M}_{t}, \boldsymbol{M}_{l_1}, \boldsymbol{M}_{l_2} \right].
\end{equation}

Force and moment equilibrium relationships yield:
\begin{equation}
\label{eq24}
    \sum_{i \in \{p,u_1,u_2,t,l_1,l_2\}} \boldsymbol{F}_i +m_u \boldsymbol{a}_u = \sum_{j \in \{x,y,z\}} \boldsymbol{F}_j,
\end{equation}
\begin{equation}
\label{eq25}
    \sum_{i \in \{p,u_1,u_2,t,l_1,l_2\}} \boldsymbol{M}_i+I_u \boldsymbol{\beta}_u = \sum_{j \in \{x,y,z\}} \boldsymbol{M}_j,
\end{equation}
where $m_u$ and $I_u$ are mass and moment of inertia about $CG$ of the unsprung mass, $\boldsymbol{a}_u$ and $\boldsymbol{\beta}_u$ are linear and angular acceleration of the unsprung mass.

The front suspension has 2 DOF: steering and bumping. 
We denote the steering rack displacement as $x_a$ and the expansion/contraction of the spring as $x_d$, both of which are measurable by sensors. 
These parameters directly influence suspension geometry, so hard point coordinates are expressed as functions of $x_a$ and $x_d$ (e.g., $\overrightarrow{\bold{u_1}(x_a,x_d)}$, $\overrightarrow{\bold{u_2}(x_a,x_d)}$), which are solved by rotation-sphere-sphere-rotation (RSSR) models in \cite{zeng2024analysis}.

By installing sensors on the spring-damper, we calculate $\boldsymbol{F}_p$ by $x_d$ and $\dot{x}_d$. 
Furthermore, tire forces $\boldsymbol{F}_x$ and $\boldsymbol{F}_y$ are related to $\boldsymbol{F}_z$ according to the magic formula \cite{pacejka2005tire}. 
Therefore, $\boldsymbol{F}_z$ is solved using $x_a$, $x_d$, $\dot{x_d}$, and $\boldsymbol{a}_u$ as:
\begin{equation}
\boldsymbol{F}_z=f(\boldsymbol{x})=f([x_a,x_d,\dot{x}_d,\boldsymbol{a}_u]^\top),
\end{equation}
where $f$ represents all the solving processes from Equation (\ref{eq2}) to Equation (\ref{eq25}).
Then, the complex suspension dynamics will be integrated into the BNN as physical guidance.

\subsection{Bayesian Physics-Informed Neural Network}
\subsubsection{Physics-Informed Bayesian Framework}
PINNs are a class of neural networks designed to solve complex physical systems by embedding the governing physical laws directly into the network's loss function \cite{TrainingPINNs}. 
This integration of a priori physical knowledge helps regularize the model, mitigating overfitting and preventing non-physical solutions \cite{Karniadakis2021}. 
Therefore, we use a general form for physical constraints:
\begin{equation}
\label{e1}
    \mathcal{F}_{\text{phy}}[u_{\boldsymbol{w}}(\boldsymbol{x}), \boldsymbol{x}] = 0, \quad \boldsymbol{x} \in \Omega, \\
\end{equation}
and enforce consistency with the available observations $\mathcal{O}$:
\begin{equation}
    \mathcal{B}[u_{\boldsymbol{w}}(\boldsymbol{x}), \boldsymbol{x}] = 0, \quad \text{for each data pair } (\boldsymbol{x}, u) \in \mathcal{O},
\end{equation}
where $u_{\boldsymbol{w}}(\boldsymbol{x})$ is the neural network solution with parameters (weights and biases) $\boldsymbol{w}$, which approximates the true physical relationship. 
The input $\boldsymbol{x}$ belongs to the operational domain $\Omega \subseteq \mathbb{R}^d$, and $d$ is the dimension of $\boldsymbol{x}$. 
$\mathcal{F}_{\text{phy}}$ is the physics-based residual operator derived from the system's governing equations, and $\mathcal{B}$ is the constraint operator that penalizes mismatches with observations.

However, conventional PINNs provide only point estimates for the weights $\boldsymbol{w}$, failing to capture uncertainties inherent in noisy, real-world data \cite{Li2023}. 
To address this, we adopt a Bayesian approach. 
Instead of finding a single optimal $\boldsymbol{w}$, BNNs aim to infer the posterior probability distribution of the parameters given a set of observations $\mathcal{O}$, denoted as $p(\boldsymbol{w} | \mathcal{O})$ \cite{B-PINNs}.
According to Bayes' theorem, this posterior is:
\begin{equation}
\label{e4}
p(\boldsymbol{w} | \mathcal{O}) = \frac{p(\mathcal{O} | \boldsymbol{w}) p(\boldsymbol{w})}{p(\mathcal{O})},
\end{equation}
where $p(\boldsymbol{w})$ is the prior distribution over the weights, $p(\mathcal{O} | \boldsymbol{w})$ is the likelihood of the observations given the weights, and $p(\mathcal{O})$ is the marginal likelihood or evidence.

With the posterior distribution of the weights, we compute the predictive posterior distribution for a new input $\boldsymbol{x}$, which provides not only a prediction but also a measure of its uncertainty. 
However, the calculation of the posterior $p(\boldsymbol{w} | \mathcal{O})$ is typically analytically intractable, especially for DNNs \cite{Guan2024}. 
Therefore, to obtain the posterior prediction for an input $\boldsymbol{x}$, we first need to draw $N$ samples, $\{\boldsymbol{w}^{(i)}\}_{i=1}^{N}$, from the posterior distribution $p(\boldsymbol{w}|\mathcal{O})$. 
The statistics of the resulting output samples, $\{u(\boldsymbol{x}; \boldsymbol{w}^{(i)})\}_{i=1}^{N}$, such as the mean and standard deviation, can then be used to represent the final prediction and its associated uncertainty, respectively \cite{EKI-B-PINNs}. 



\subsubsection{Variational Inference with Physics-Informed Likelihood}

To approximate the true posterior $p(\boldsymbol{w}|\mathcal{O})$, we employ variational inference (VI). 
VI introduces a simpler, tractable distribution $q_{\boldsymbol{\zeta}}(\boldsymbol{w})$, parameterized by variational parameters $\boldsymbol{\zeta}$, to approximate the true posterior. 
The goal is to tune $\boldsymbol{\zeta}$ by minimizing the Kullback-Leibler (KL) divergence between the approximate and true posteriors.

Following the PINN framework, the ``observations" $\mathcal{O}$ implicitly contain two types of information: the measured data and the physical laws. 
Therefore, we decompose the log-likelihood term $\log p(\mathcal{O} | \boldsymbol{w})$ into two parts: a data-matching term and a physics-residual term.

First, the likelihood of observing the measurement data is given by:
\begin{equation}
    p(\mathcal{O}_{\text{data}} | \boldsymbol{w}) = \prod_{(\boldsymbol{x}_i, u_i) \in \mathcal{O}} \mathcal{N}(u_i | u_{\boldsymbol{w}}(\boldsymbol{x}_i), \sigma_{\text{data}}^2),
\end{equation}
where we assume the measurement noise is Gaussian with variance $\sigma_{\text{data}}^2$ \cite{B-PINNs}.

Second, the physical constraint is enforced at a set of collocation points $\{\boldsymbol{x}_c\}_c \subset \Omega$. 
Thus, the likelihood is:
\begin{equation}
    p(\mathcal{O}_{\text{phy}} | \boldsymbol{w}) = \prod_{\boldsymbol{x}_c} \mathcal{N}(0 | \mathcal{F}_{\text{phy}}[u_{\boldsymbol{w}}(\boldsymbol{x}_c), \boldsymbol{x}_c], \sigma_{\text{phy}}^2),
\end{equation}
where $\sigma_{\text{phy}}^2$ is interpreted as a precision parameter for enforcing the physical law.

Combining these terms, the objective is to minimize the negative evidence lower bound (ELBO) \cite{blundell2015weight}, which is equivalent to maximizing ELBO itself. 
The loss function is:
\begin{equation}
\label{eq_loss}
\begin{aligned}
\mathcal{L}(\boldsymbol{\zeta}) &= \mathbb{E}_{\boldsymbol{w} \sim q_{\boldsymbol{\zeta}}} [-\log p(\mathcal{O}_{\text{data}} | \boldsymbol{w}) - \log p(\mathcal{O}_{\text{phy}} | \boldsymbol{w})] \\&+ \text{KL}(q_{\boldsymbol{\zeta}}(\boldsymbol{w}) \, || \, p(\boldsymbol{w})), \\
\end{aligned}
\end{equation}
where the first term represents the data-mismatch loss and the physics-residual loss, respectively, weighted by their corresponding precision. 
The final KL term acts as a regularizer, encouraging the approximate posterior to stay close to the prior distribution over the weights.

\begin{figure}[ht!]
    \centering
    \vspace{6pt}
    \includegraphics[width=1\linewidth]{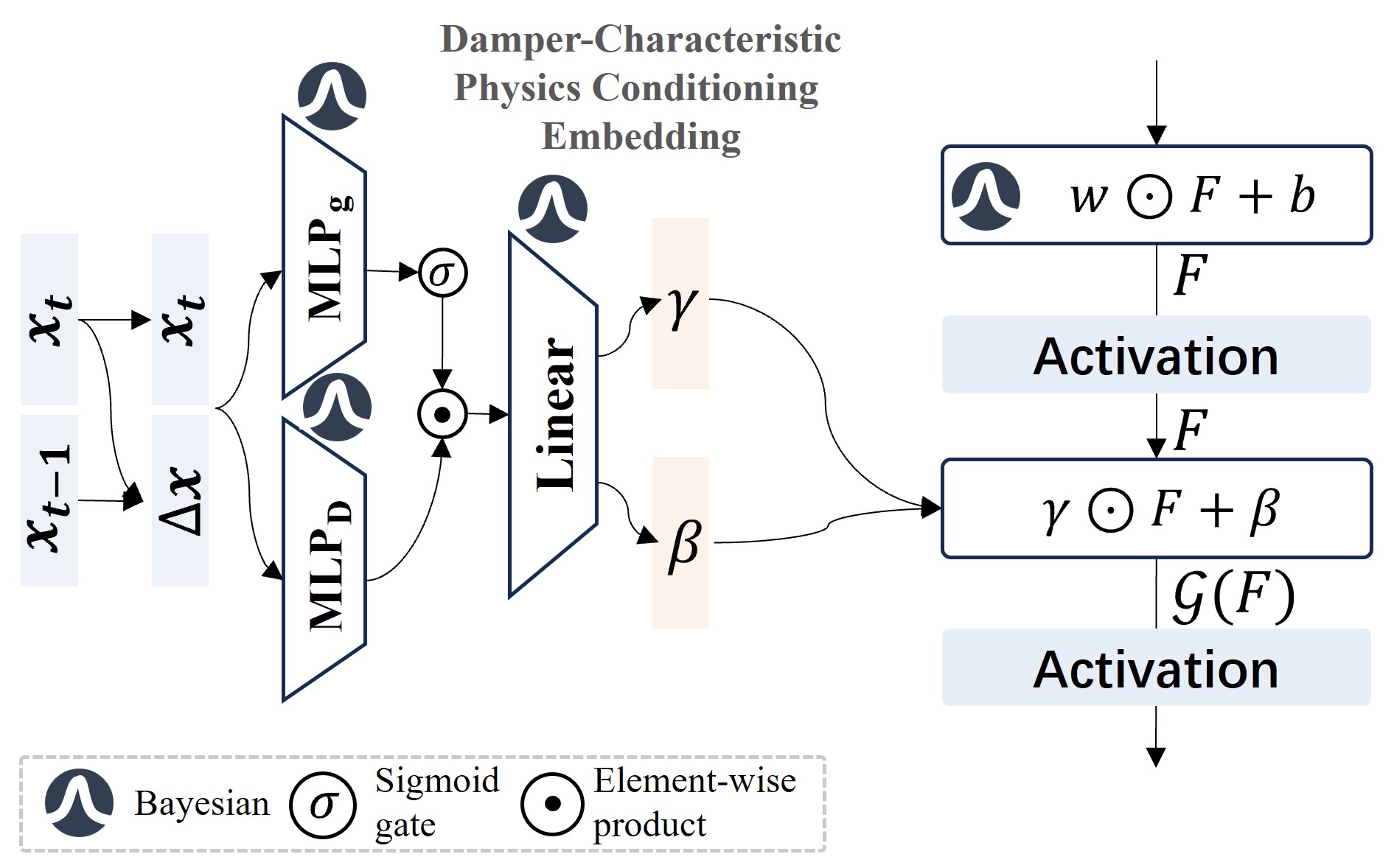}
    \caption{Structure of the proposed DPC module. }
    \label{fig:damper}
\end{figure}


\subsubsection{Normal-Sigmoid-Dropout Method}

Recent works \cite{Zhang2019,Dropout-Bayesian} have shown that training a neural network with dropout is equivalent to a form of variational inference. 
However, standard dropout, where neurons are randomly set to zero, can lead to poor convergence in complex systems. 
To address this, we propose a normal-sigmoid-dropout (NS-Dropout) method. 
For a given layer's neuron activations $x_{n+1}$, we introduce a stochastic noise matrix $H$. 
Its elements $h_i$ are sampled from a normal distribution, $h_i \sim \mathcal{N}(0, \sigma^2)$, where $\sigma$ is a tunable hyperparameter. 
The noise matrix is:
\begin{equation}
 H = \frac{1}{2}\phi(H) + \frac{\boldsymbol{I}}{2},
\end{equation}
where $\phi$ is the sigmoid function and $\boldsymbol{I}$ is an identity matrix. 
This transformation bounds the multiplicative noise for each neuron within the range $(\frac{1}{2}, 1)$, preventing the complete deactivation of neurons. 
The dropout-applied neuron values $x'_{n+1}$ are then computed as:
\begin{equation}
 x'_{n+1} = H \odot x_{n+1},
\end{equation}
where $\odot$ denotes the element-wise product. 
This method provides a more stable way to inject stochasticity, facilitating robust variational inference.

\subsection{Damper-Characteristic Physics Conditioning Embedding}

To deeply embed the physical dynamics into our model, we propose a novel physics-informed conditioning encoder, termed the DPC module, which is inspired by the behavior of a mechanical damper and its purpose is to generate dynamic conditioning parameters for the main BNN, which is shown in Figure \ref{fig:damper}.
The proposed DPC module leverages the FiLM technique \cite{perez2018film}, which applies a feature-wise affine transformation $\mathcal{G}(F) = \gamma \odot F + \beta$ to the intermediate activation vector $F$ of a network.

The DPC encoder is designed to produce these modulation parameters, i.e. $\gamma$ and $\beta$, from the vehicle's real-time state dynamics. The encoder takes a time-series of sensor inputs, specifically the current state $\boldsymbol{x}_t$ and the previous state $\boldsymbol{x}_{t-1}$ as its input. Its operation begins by explicitly modeling the velocity-dependent force of the damper. It estimates the discrete-time ``velocity" of the state by calculating the difference between consecutive time steps as $\Delta \boldsymbol{x}_t = \boldsymbol{x}_t - \boldsymbol{x}_{t-1}$.

Following this, the encoder utilizes two parallel pathways to learn a sophisticated, state-dependent damping representation. The first pathway learns a nonlinear $D_t$ by processing both the current state and its rate of change through a multi-layer perceptron (MLP):
\begin{equation}
    D_t = \text{MLP}_D([\Delta \boldsymbol{x}_t, \boldsymbol{x}_t]).
\end{equation}

Simultaneously, the second pathway processes only the current state $\boldsymbol{x}_t$ to generate a gating signal $g_t$, which is constrained between 0 and 1 by a sigmoid function $\sigma$:
\begin{equation}
    g_t = \sigma(\text{MLP}_g(\boldsymbol{x}_t)).
\end{equation}

This gate acts as a dynamic control valve, allowing the model to learn how much of the damping effect to apply based on the vehicle's current state.
The final output is:
\begin{equation}
    [\gamma_t, \beta_t] = \text{Linear}(g_t \odot D_t).
\end{equation}

These generated parameters $\gamma_t$ and $\beta_t$ are then injected into the main Damper-B-PINN. This allows the physical dynamics encoded by the DPC module to directly guide the feature extraction process throughout the entire network.

\section{Experiments and Results}
\label{experiment}

\subsection{Experimental Setup}

\begin{figure*}[ht]
    \centering
    \vspace{6pt}
    \includegraphics[width=1\linewidth]{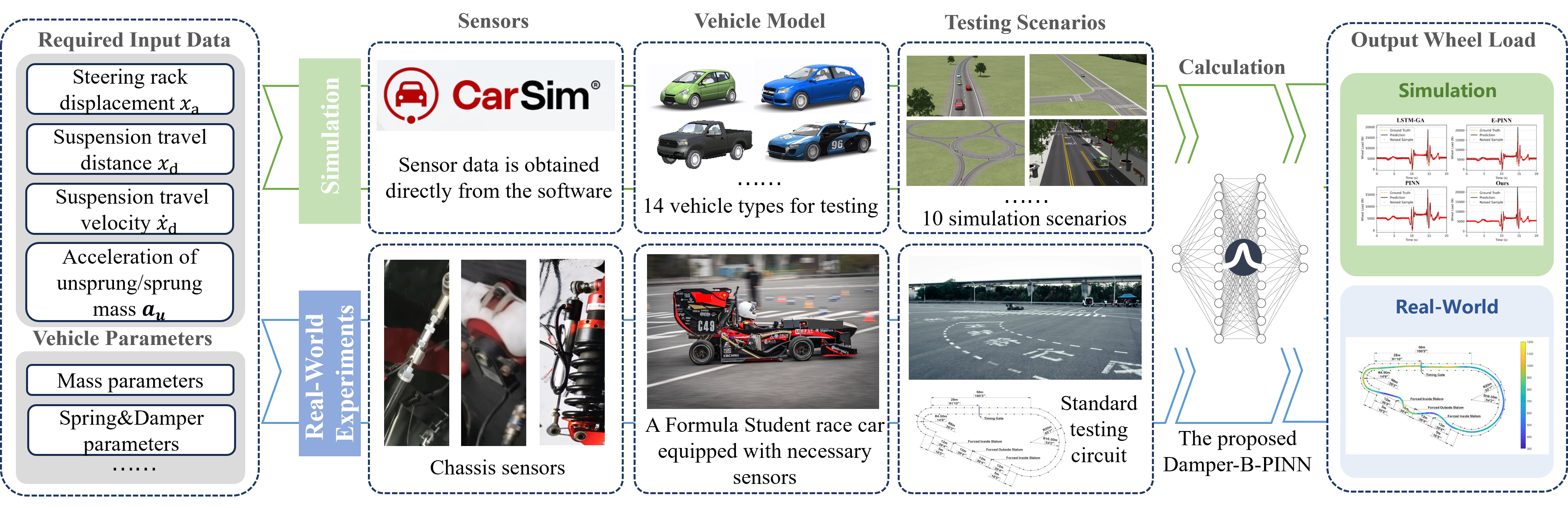}
    \caption{Overview of the experiments: The input data are directly obtained by software in the simulation, while in the real-world experiments, they are collected by the corresponding sensors on the chassis. In the simulation, 14 types of vehicle models and 10 types of scenarios are selected for data collection. The real-world experiments are conducted on a Formula Student race car and completed in a standard test site. The collected data undergo offline calculation to obtain the output.}
    \label{fig:exp}
\end{figure*}

The simulations are performed in CarSim 2019.0 platform, and the real-world tests are conducted on a Formula Student race car shown in Figure \ref{fig:race}.
Compared with passenger cars, race cars operate under more extreme working conditions.
The overview of the experiments is shown in Figure \ref{fig:exp}.

\subsection{Comparative Experiments}

\subsubsection{Simulation Results}

We select 10 simulation scenarios, including two different driving styles: smooth driving and aggressive driving, and test with 14 vehicle types. 
For each vehicle type, several 20-second data segments are collected under each scenario. 
Training set accounts for 70\% of total.
Sensor data required by the algorithm are directly obtained from the software, and the ground truth of wheel loads is also directly obtained.
We choose LSTM-GA \cite{zeng2024wheel}, E-PINN \cite{Guan2024}, and PINN \cite{Long2022} as comparative baseline methods, with \textit{RMSE} (root mean square error) and \textit{MaxError} (mean of max error of each segment) as evaluation metrics. 
The quantitative results are shown in Table \ref{simu}.

\begin{table}[ht]
\vspace{6pt}
\caption{Quantitative simulation results.}
\label{simu}
\setlength{\tabcolsep}{2.5pt} 
\begin{center}

\begin{tabular}{l|cc|cc}
\toprule
\multirow{2}*{\textbf{Method}} & \multicolumn{2}{c|}{\textbf{Smooth Driving}} & \multicolumn{2}{c}{\textbf{Aggressive Driving}}\\
~&{RMSE(N)$\downarrow$} & {MaxError(N)$\downarrow$} & {RMSE(N)$\downarrow$} & {MaxError(N)$\downarrow$}\\
\midrule

LSTM-GA \cite{zeng2024wheel} &  582.840 &   1089.432& 1150.434&1553.917\\
E-PINN \cite{Guan2024}  &  979.510 & 1307.613& 1294.784&1767.661\\
PINN \cite{Long2022}&  598.617 &  \textbf{843.677}& 1002.531&1389.431\\
\textbf{Ours}     &  \textbf{550.397} &   850.431& \textbf{975.071}& \textbf{1103.391}\\

\bottomrule
\end{tabular}

\end{center}

\end{table}

\begin{figure}[ht!]
\centering
\vspace{6pt}
\includegraphics[width=1\linewidth]{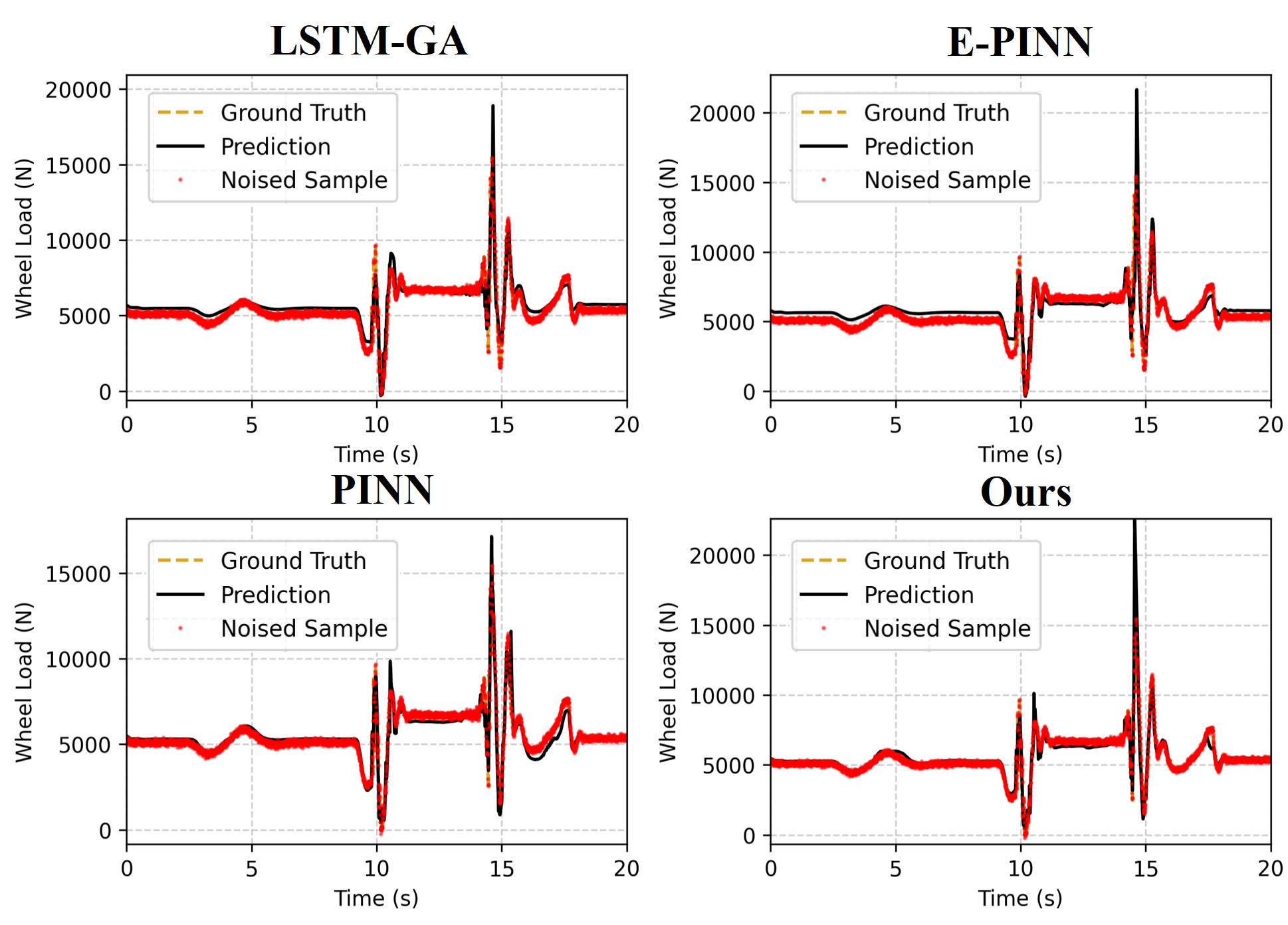}
\caption{Simulation results on a segment of aggressive driving.}
\label{fig:simu1}
\end{figure}

\begin{figure}[ht!]
\centering
\vspace{6pt}
\includegraphics[width=1\linewidth]{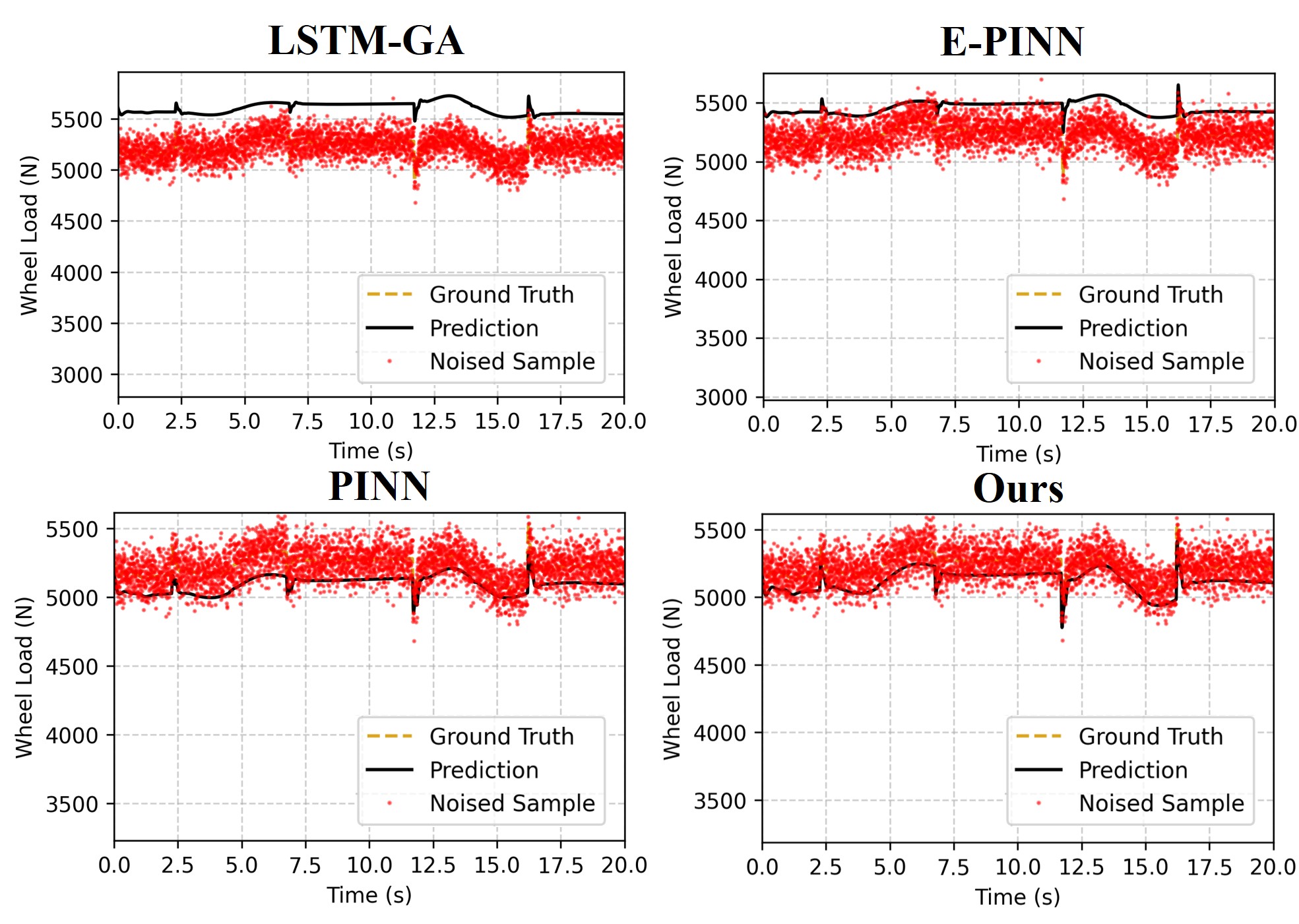}
\caption{Simulation results on a segment of smooth driving.}
\label{fig:simu2}
\end{figure}

We select two sets of data for demonstration, one for aggressive driving and another for smooth driving, as shown in Figure \ref{fig:simu1} and \ref{fig:simu2}. The yellow dashed line represents the ground truth, and the red dots indicate the noise simulated by the software.
According to the simulation results, our Damper-B-PINN method achieves the best performance. 
Under aggressive driving conditions, it is able to capture the details and patterns of the data even when wheel loads change drastically. 
However, in some cases, the \textit{MaxError} indicator of our method does not reach the optimal level. 
This is because Bayesian sampling in certain extreme regions may introduce significant variance, which can lead to data drift. 
Unlike other methods that do not involve Bayesian sampling, this results in the spike-like numerical anomaly at the $14.5$-second mark in Figure \ref{fig:simu1}. 
However, the introduction of the Bayesian method has improved the calculation accuracy in other noise-containing regions, and the overall average \textit{MaxError} remains within a controllable range.
Under the smooth driving condition, our method also achieves the best results. Guided by the physical model, it is able to accurately extract tiny wheel load fluctuations, such as those at $12$ seconds and $16$ seconds as shown in Figure \ref{fig:simu2}.

\begin{figure*}[ht]
    \centering
    \vspace{6pt}
    \includegraphics[width=1\linewidth]{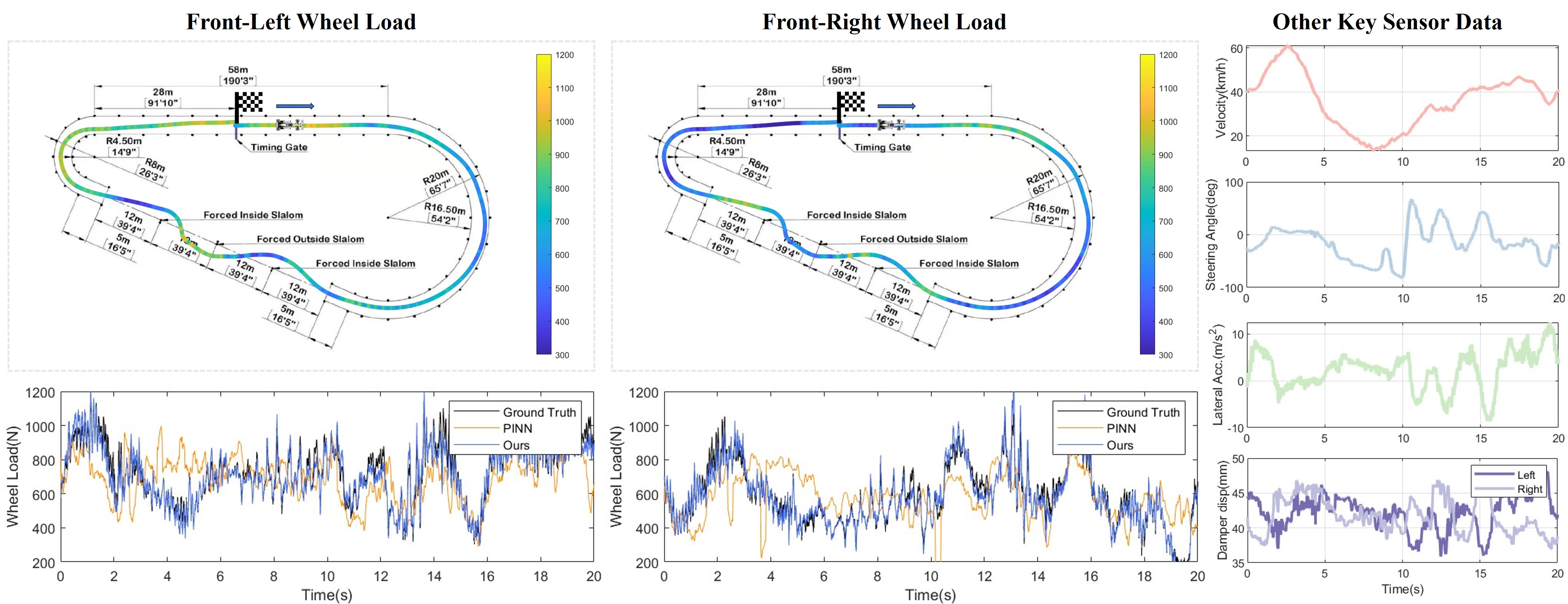}
    \caption{Real-world test results extracted from one lap of the race track: The distributions of front-left and front-right wheel loads are shown in the top-left two figures; the bottom-left two figures compare the wheel loads measured by our \textbf{Damper-B-PINN} method with those from PINNs; and the right side displays data from vehicle velocity sensor, steering angle sensor, acceleration sensor and damper displacement sensor.
}
    \label{fig:res}
\end{figure*}

\subsubsection{Real-World Test Results}

Real-world tests are conducted on a standard race track, including various scenarios such as long straights, high-speed corners, low-speed corners, and slaloms, shown in Figure \ref{fig:res}. 
Considering the lightweight requirements of racing cars, the computing power of on-board hardware is limited. 
Therefore, after collecting data through sensors, we perform offline calculations. 
The ground truth of wheel load is collected by a set of wheel-load transducer installed outside of the wheel rim.
We select PINN \cite{Long2022} as the comparative method, also with \textit{RMSE} and \textit{MaxError} as evaluation metrics.
The quantitative comparison results of the real-world experiments are shown in Table \ref{real_res}.
It should be noted that since the mass of the race car is very light, with a total weight of only about 280kg including the driver, the wheel load and error values will be smaller than those of passenger cars and commercial vehicles in Carsim simulations.

\begin{table}[htbp!]
\vspace{6pt}
\caption{Quantitative real-world test results.}
\label{real_res}

\begin{center}

\begin{tabular}{l|c|c}
\toprule
\textbf{Method} & {RMSE(N)$\downarrow$} & {MaxError(N)$\downarrow$} \\
\midrule
PINN \cite{Long2022}   & 479.563 &  1618.936\\
\textbf{Ours}  &  \textbf{255.049} &   \textbf{1282.934}\\
\bottomrule
\end{tabular}
\end{center}
\end{table}


We take experimental results of one lap for demonstration in Figure \ref{fig:res}. 
It can be observed that in real-world tests, due to the influence of sensor noise and the asymmetric distribution of the vehicle's static load, the traditional PINN method fails to recognize the noise in the vehicle model, and it incorrectly assumes a symmetric distribution of the vehicle's mass, leading to deviations in load calculation. 
In contrast, since the static load of an actual vehicle cannot be symmetrically distributed, our Damper-B-PINN method, which relies on refined suspension modeling and Bayesian processing of noise, achieves more favorable results.
Real-world experiments fully demonstrate the stability of our Damper-B-PINN method. 
It can be observed that high accuracy is maintained even when the lateral acceleration reaches nearly 10 m/s$^2$. 
Our system provides a robust data source for scenarios including racing car control, driving assistance in extreme conditions, and vehicle risk avoidance, which is of significant importance.

\subsection{Ablation Studies}

We conduct ablation experiments on several key modules, and the results are shown in Table \ref{ablation}.
It is noted here that ``Basic model" refers to establishing a model for the entire vehicle \cite{sieberg2019hybrid} instead of modeling the suspension linkage.

\begin{table}[ht]
\vspace{6pt}
\caption{Ablation studies.}
\label{ablation}
\setlength{\tabcolsep}{1pt} 
\begin{center}

\begin{tabular}{l|cc|cc}
\toprule
\multirow{2}*{\textbf{Method}} & \multicolumn{2}{c|}{\textbf{Simulation}} &  \multicolumn{2}{c}{\textbf{Real-World Experiment}}\\
~ & {RMSE(N)$\downarrow$} & {MaxError(N)$\downarrow$} & {RMSE(N)$\downarrow$}& {MaxError(N)$\downarrow$}\\
\midrule
Basic Model  &  894.153 &   1187.390& 560.697& 1528.653\\
w/o Bayesian   & 819.433 &  1236.941& 381.713& 1339.161\\
w/o Physics(DPC) &  770.003 &   1159.105& 360.988& 1597.554\\
w/o NS-Dropout   &  759.641 &   1009.573& 263.793& 1729.601\\
\midrule
\textbf{Full}   &  \textbf{700.414} &  \textbf{976.911}& \textbf{255.049}& \textbf{1282.934}\\

\bottomrule
\end{tabular}

\end{center}

\end{table}

When a Basic Model without suspension modeling is selected, due to the highly nonlinearity of vehicle system, it will lead to significant deviations through relying solely on the overall parameters of vehicle body. 
In addition, the guidance of the physical loss ensures that the output of the neural network meets the basic constraints and will not produce an excessively large \textit{MaxError}. The denoising function of the Bayesian operation has also been proven to be effective, while NS-Dropout further improves the network performance.



\section{Conclusion}
\label{concluion}

This paper proposes a framework for measuring the dynamic wheel load of vehicles. 
Firstly, through refined modeling of the suspension linkage system, a nonlinear instantaneous dynamic system is established to characterize the complex geometric structure of the suspension. 
Subsequently, the chassis dynamics is integrated into the neural network as physical information by the DPC module, and Bayesian operations are adopted to enhance the network's robustness against noisy inputs. 
We conduct experiments using CarSim simulation software and a Formula Student race car, verifying the effectiveness and stability of our method under various working conditions. 
However, considering the requirements for race car light-weighting and the constraints on on-board computing power, we have not yet realized real-time on-board computation. 
In addition, due to the need for special design of relevant chassis sensors, we have only conducted tests on the Formula Student race car and have not verified the method on passenger vehicles. 
Future work will attempt to solve the above issues.
	
	
	\bibliographystyle{IEEEtran}
	\bibliography{root} 
	
\end{document}